\documentclass[a4paper]{fairmeta}
\usepackage{fancyvrb}

\usepackage{algorithm}
\usepackage{algorithmicx}
\usepackage{algpseudocode}
\usepackage{float}
\newfloat{algorithm}{t}{lop}

\makeatletter
\newcommand\notsotiny{\@setfontsize\notsotiny{5.5}{8}}
\makeatother

\definecolor{lightorange}{RGB}{255,220,190}
\definecolor{lightblue}{RGB}{190,190,255}
\definecolor{lightgray}{RGB}{224,224,224}
\definecolor{lightgreen}{RGB}{190,255,190}

\makeatletter
\pgfdeclareshape{with q kv}{
  \inheritsavedanchors[from=rectangle]
  \inheritanchorborder[from=rectangle]
  \inheritanchor[from=rectangle]{center}
  \inheritanchor[from=rectangle]{north}
  \inheritanchor[from=rectangle]{south}
  \inheritanchor[from=rectangle]{west}
  \inheritanchor[from=rectangle]{east}
  \inheritanchor[from=rectangle]{north west}
  \inheritanchor[from=rectangle]{north east}
  \inheritanchor[from=rectangle]{south west}
  \inheritanchor[from=rectangle]{south east}

  \anchor{input q}{
    \southwest
    \pgf@xa=\pgf@x
    \pgf@ya=\pgf@y
    \northeast
    \pgf@xb=\pgf@x
    \pgf@yb=\pgf@y
    \pgf@x=\pgf@xa
    \advance\pgf@x by  0.25\pgf@xb
    \advance\pgf@x by -0.25\pgf@xa
    \pgf@y=\pgf@ya
  }

  \anchor{input kv}{
    \southwest
    \pgf@xa=\pgf@x
    \pgf@ya=\pgf@y
    \northeast
    \pgf@xb=\pgf@x
    \pgf@yb=\pgf@y
    \pgf@x=\pgf@xa
    \advance\pgf@x by  0.75\pgf@xb
    \advance\pgf@x by -0.75\pgf@xa
    \pgf@y=\pgf@ya
  }

  \backgroundpath{
    \pgfpathrectanglecorners{\southwest}{\northeast}
  }
}
\makeatother

\title{The Free Transformer}

\author[1]{Fran\c{c}ois Fleuret}

\affiliation[1]{FAIR at Meta}

\abstract{%
We propose an extension of the decoder Transformer that conditions its
generative process on random latent variables which are learned
without supervision thanks to a variational procedure. Experimental
evaluations show that allowing such a conditioning translates into
substantial improvements on downstream tasks.
}

\date{\today}
\correspondence{\email{fleuret@meta.com}}

\usepackage{xifthen}
\usepackage[utf8]{inputenc}
\usepackage{multirow,xspace}
\usepackage{amsmath,amssymb,dsfont}

\usepackage[round]{natbib}
\usepackage{tikz}
\usetikzlibrary{math}
\usetikzlibrary{arrows,arrows.meta,calc}
\usetikzlibrary{patterns,backgrounds}
\usetikzlibrary{positioning,fit}
\usetikzlibrary{shapes.geometric,shapes.multipart}
\usetikzlibrary{patterns.meta,decorations.pathreplacing,calligraphy}
\usetikzlibrary{tikzmark}
\usetikzlibrary{decorations.pathreplacing}
\usetikzlibrary{decorations.pathmorphing}

\def\kl{\mathds{D}_{\text{KL}}}



\newcounter{nbdrafts}
\makeatletter
\newcommand{\checknbdrafts}{
\ifnum \thenbdrafts > 0
\@latex@warning@no@line{*WARNING* The document contains \thenbdrafts \space draft note(s)}
\fi}

\makeatother

\newcommand{\modelname}{{Free Transformer}\xspace}

\begin{document}

\maketitle


  %



  %

\section{Introduction}
%
%

Since their invention, the Transformer \citep{transformer_2017}, and
more specifically the decoder-only Transformers used originally for
the GPT series of models \citep{gpt_2018}, have become the core
components of AI systems.

It is remarkable that, after almost a decade, and in spite of
improvements on many aspects of this class of methods, the
autoregressive modelling of Transformers remains essentially
unchallenged. We propose in this paper to revisit this key design
aspect by allowing richer and more natural density models to emerge:
\begin{itemize}

\item We extend the auto-regressive model of the decoder Transformer
  by allowing the conditioning on latent variables, thanks to a
  formulation as a conditional Variational Autoencoder
  (\S~\ref{sec:cond-vari-auto}).

\item We propose an implementation that requires a very modest
  computational and memory usage overhead
  (\S~\ref{sec:model-structure}).

\end{itemize}

The benefits of the proposed method are shown by training 1.5B and 8B
models from scratch and assessing performance on multiple downstream
benchmarks (\S~\ref{sec:experiments}).

\section{Motivation}

Decoder Transformers are auto-regressive discrete density
approximators. They model a sequence of tokens $S_1, \dots, S_T$ by
estimating the conditional distribution of each given those preceding
it. Sampling is done by generating one token after another, each time
computing the distribution of the next symbol given those generated so
far.

The only density modelling and sampling that such models implement is
that of the generated tokens. In particular, a decoder Transformer
does not make additional latent decisions about the stream of symbols
to generate. Its only decisions are the choices of the tokens
themselves.

Consider, for instance, that we train such a model to generate movie
reviews and that we want to have two clearly separated categories of
negative and positive reviews. Given a large enough model and the
necessary amount of training data, there is no doubt that a decoder
Transformer trained on a dataset of that form would work perfectly and
would generate these two types of reviews.  However, to do so, it
would generate tokens one after another and decide, based on the words
generated so far, whether the review it is currently generating is a
positive or a negative one, and continue the process accordingly.  In
particular, \emph{the model would not make the explicit decision to
generate a negative or a positive review}. It would produce tokens,
and this notion of a negative or positive review would be implicit in
their posterior probabilities.

Due to the chain rule, any density can be modelled as
autoregressive. However, in particular when the ``natural'' structure
involves conditioning on latent variables, the autoregressive model
of the signal may be a great deal more complex than the full joint
model including the latent.

We can consider a simple example illustrating that point. Let $Z \sim
\mathcal{B}(0.5)$ be a latent ``coin flip'', and $X_1, \dots, X_T$ be
equal to $Z$ with independent flips of probability $\epsilon$.

The $X_t$s are conditionally independent given $Z$, and we have
\begin{equation}
P(X_{t+1} = 1 \mid Z=z) = \epsilon z + (1-\epsilon) (1-z)
\end{equation}
however, expressed as an auto-regressive model without $Z$, we get:
\begin{equation}
P(X_{t+1} = 1 \mid X_1=x_1, \dots, X_t=x_t) = \frac{\left(\frac{\epsilon}{1-\epsilon}\right)^{{\sum_{s=1}^t x_s}} (1-\epsilon)^{t+1} +  \left(\frac{1-\epsilon}{\epsilon}\right)^{{\sum_{s=1}^t x_s}} \epsilon^{t+1}}{\left(\frac{\epsilon}{1-\epsilon}\right)^{{\sum_{s=1}^t x_s}} (1-\epsilon)^{t} +  \left(\frac{1-\epsilon}{\epsilon}\right)^{{\sum_{s=1}^t x_s}} \epsilon^{t}}.
\end{equation}

We could easily come with worse examples when expressed
autoregressively, for instance when the latent variables are positions
in the sequence, e.g. the index where a certain pattern occurs as in
the example of \S~\ref{sec:synthetic-dataset}. What we observe in such
cases is that it requires running estimates of probabilities
(``probability that the target appears here'') for which estimation
errors are unavoidable and problematic.

The consequence is that a purely auto-regressive density model suffers
potentially from several drawbacks:
\begin{itemize}

\item It requires an unnecessarily complicated computation, and
  greater capacity, to implicitly make post-hoc decisions or infer
  latent quantities from the generated tokens.

\item It may be sent off track during the process if, by mistake, a
  few tokens generated are erroneous, ambiguous or contradictory with
  those generated previously.

\item Key concepts do not appear spontaneously due to the "natural"
  factorization of the distribution, but are built post-hoc by
  necessity to fit the training samples better. This may be a
  fundamental weakness when operating out of distribution.

\end{itemize}

The main objective of the present work is to address these issues by
providing the model with the freedom of conditioning its
auto-regressive process on latent random quantities that are not
imposed by the training examples.

For instance, for the review generator example above, the model could
use a random Boolean value to decide once for all whether the tokens
it produces are from the distribution of negative or positive reviews,
removing the need for a complicated posterior estimate from the tokens
already generated.


\section{Method}


Any latent random value $Y_r$, whatever its statistical dependency
with the tokens $S_1, \dots, S_t$ and other latent $Y_1, \dots,
Y_{r-1}$ sampled so far, can be expressed under reasonable assumptions
as $f_r(S_1, \dots, S_t, Y_1, \dots, Y_{r-1}, Z_r)$ where $Z_r$ is a
value coming from a random generator.

Hence, if we provide the model with enough random values $Z_1, Z_2,
\dots$ sampled independently during generation, a proper training
procedure could in principle build families of latent variables with
arbitrary dependency structure, as long as the model's capacity allows
it to encode $f_r$.


In the same way that the choice of a token during sampling can be
expressed as a function of a random value and the logits, any
activation which is a function of a random value and other activations
can be interpreted as a decision made by the model during the
generative process. Such decisions make the latent activation
non-deterministic functions of the tokens, and observing the latter
only gives a partial information about the former.


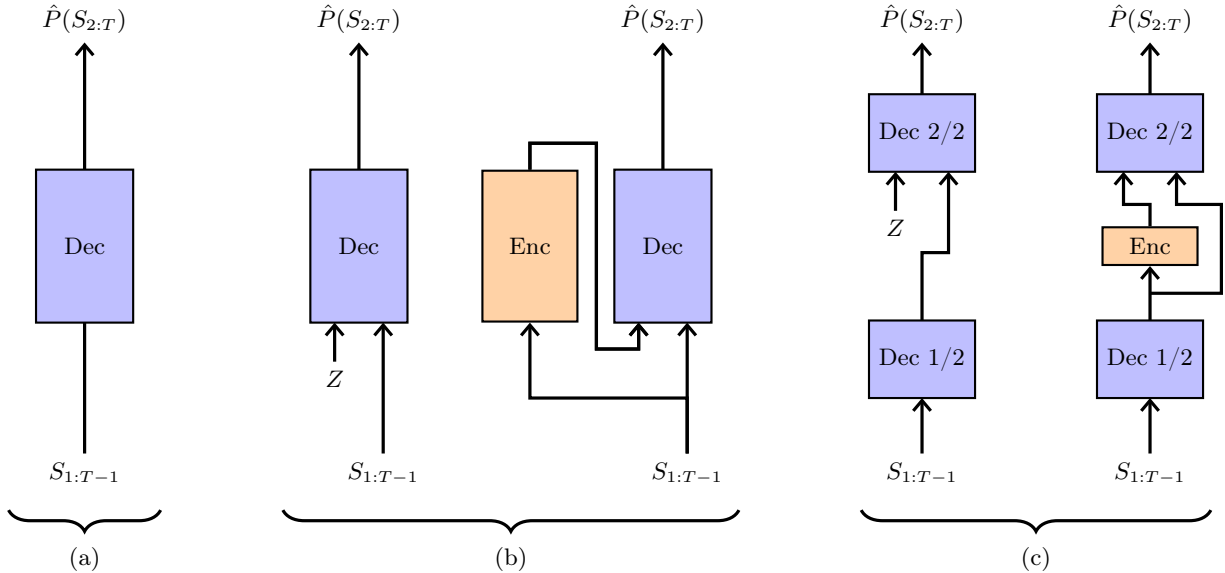
\begin{figure} 

%
\begin{tikzpicture}[
    font=\small,    
    >={Straight Barb[angle'=80,scale=1]},
    dec/.style={draw=black,thick,fill=blue!25,minimum width=1.25cm},
    enc/.style={draw=black,thick,fill=orange!35,minimum width=1.25cm},
    qkv/.style={with q kv,text height=3.5ex,text depth=3.5ex,
      label={[anchor=south,yshift=-1]input q:\tiny Q},
      label={[anchor=south]input kv:\tiny KV},
    },
    flow/.style={very thick},
  ]

\draw[-,decoration={brace,amplitude=8pt,mirror},very thick,decorate] (-1, -3.5) -- ++(2, 0) node[midway,below,yshift=-10pt] {(a)};

\node[dec,minimum height=2cm,with q kv] (decoder) at (0, 0) {Dec};

\node (input) at ($(decoder.south)+(0, -2)$) {$S_{1:T-1}$};
\node (output) at ($(decoder.north)+(0, 2)$) {$\hat{P}(S_{2:T})$};

\draw[flow] (input) -- (decoder);
\draw[flow,->] (decoder) -- (output);
\end{tikzpicture}
\hspace*{\stretch{1}} 
\begin{tikzpicture}[
    font=\small,    
    >={Straight Barb[angle'=80,scale=1]},
    dec/.style={draw=black,thick,fill=blue!25,minimum width=1.25cm},
    enc/.style={draw=black,thick,fill=orange!35,minimum width=1.25cm},
    qkv/.style={with q kv,text height=3.5ex,text depth=3.5ex,
      label={[anchor=south,yshift=-1]input q:\tiny Q},
      label={[anchor=south]input kv:\tiny KV},
    },
    flow/.style={very thick},
  ]


\draw[-,decoration={brace,amplitude=8pt,mirror},very thick,decorate] (-1, -3.5) -- ++(6, 0) node[midway,below,yshift=-10pt] {(b)};

\node[dec,minimum height=2cm,with q kv] (decoder) at (0, 0) {Dec};

\node (z) at ($(decoder.input q)+(0, -0.75)$) {$Z$};
\node (input) at ($(decoder.input kv)+(0, -2)$) {$S_{1:T-1}$};
\node (output) at ($(decoder.north)+(0, 2)$) {$\hat{P}(S_{2:T})$};
\draw[flow,->] (z) -- (decoder.input q);
\draw[flow,->] (input) -- (decoder.input kv);
\draw[flow,->] (decoder) -- (output);


\node[enc,minimum height=2cm] (encoder) at (2.25, 0) {Enc};
\node[dec,minimum height=2cm,with q kv] (decoder) at (4, 0) {Dec};

\coordinate (middle top) at ($(encoder.north)!0.5!(decoder.north)+(0, 0.35)$);
\coordinate (middle bottom) at ($(encoder.south)!0.5!(decoder.south)+(0, -0.35)$);
\node (input) at ($(decoder.input kv)+(0, -2)$) {$S_{1:T-1}$};
\node (output) at ($(decoder.north)+(0, 2)$) {$\hat{P}(S_{2:T})$};
\draw[flow,->] (encoder.north) |- (middle top) -- (middle bottom) -| (decoder.input q);
\draw[flow,->] (encoder.north) |- (middle top) -- (middle bottom) -| (decoder.input q);
\draw[flow,->] (input) -- ++(0, 1) -| (encoder.south);
\draw[flow,->] (input) -- (decoder.input kv);
\draw[flow,->] (decoder) -- (output);
\end{tikzpicture}
\hspace*{\stretch{1}} 
\begin{tikzpicture}[
    font=\small,    
    >={Straight Barb[angle'=80,scale=1]},
    dec/.style={draw=black,thick,fill=blue!25,minimum width=1.25cm},
    enc/.style={draw=black,thick,fill=orange!35,minimum width=1.25cm},
    qkv/.style={with q kv,text height=3.5ex,text depth=3.5ex,
      label={[anchor=south,yshift=-1]input q:\tiny Q},
      label={[anchor=south]input kv:\tiny KV},
    },
    flow/.style={very thick},
  ]


\draw[-,decoration={brace,amplitude=8pt,mirror},very thick,decorate] (-2.8, -3.5) -- ++(4.6, 0) node[midway,below,yshift=-10pt] {(c)};

\node[dec,minimum height=1cm,with q kv] (decoder bottom) at (-2, -1.5) {Dec 1/2};
\node[dec,minimum height=1cm,with q kv] (decoder top) at (-2, 1.5) {Dec 2/2};

\coordinate (after decoder bottom) at ($(decoder bottom.north)+(0, 0.9)$);
\node (input) at ($(decoder bottom.south)+(0, -1)$) {$S_{1:T-1}$};
\node (output) at ($(decoder top.north)+(0, 1)$) {$\hat{P}(S_{2:T})$};
\draw[flow,->] (input) -- (decoder bottom);
\draw[flow,->] (decoder top) -- (output);
\draw[flow,->] (decoder bottom.north) -- (after decoder bottom) -| (decoder top.input kv);
\node (z) at ($(decoder top.input q)+(0, -0.75)$) {$Z$};
\draw[flow,->] (z) -- (decoder top.input q);


\node[enc,minimum height=0.5cm] (encoder) at (1, 0) {Enc};
\node[dec,minimum height=1cm,with q kv] (decoder bottom) at (1, -1.5) {Dec 1/2};
\node[dec,minimum height=1cm,with q kv] (decoder top) at (1, 1.5) {Dec 2/2};

\coordinate (after decoder bottom) at ($(decoder bottom.north)!0.5!(encoder.south)$);
\coordinate (after encoder) at ($(encoder.north)!0.4!(decoder top.south)$);
\coordinate (right side) at ($(encoder.east)+(0.3, 0)$);
\node (input) at ($(decoder bottom.south)+(0, -1)$) {$S_{1:T-1}$};
\node (output) at ($(decoder top.north)+(0, 1)$) {$\hat{P}(S_{2:T})$};
\draw[flow,->] (input) -- (decoder bottom);
\draw[flow,->] (decoder top) -- (output);
\draw[flow,->] (decoder bottom.north) |- (after decoder bottom) -| (encoder.south);
\draw[flow,->] (encoder.north) |- (after encoder) -| (decoder top.input q);
\draw[flow,->] (after decoder bottom) -| (right side) |- (after encoder-|decoder top.input kv) -- (decoder top.input kv);

\end{tikzpicture}%
%


\caption{A standard decoder Transformer (a) can be extended to utilize
  a random state $Z$ in inference (b, left), in which case it has to
  be trained as a conditional VAE with an encoder (b, right). The Free
  Transformer (c) reduces the overhead of the encoder by having the
  random state $Z$ injected in its middle layer (c, left), and using
  for encoder during training the combination of its first half
  combined with one non-causal layer specific to the encoder (c,
  right). See Figure \ref{fig:free-transformer} for a detailed
  depiction of that architecture.}\label{fig:models}

\end{figure}


\begin{figure} 

\center


\newcommand{\thatfigspace}{3ex}

\begin{tikzpicture}[
    >={Straight Barb[angle'=80,scale=1]},
    next/.style={above=0.65cm of \tikzlastnode},
    next tight/.style={above=0.1cm of \tikzlastnode},
    box/.style={draw,thick,minimum height=0.75cm,minimum width=4.5cm,inner sep=0pt,align=center},
    op/.style={box,fill=blue!20},
    enc/.style={box,fill=orange!35},
    val/.style={box,fill=white},
    qkv/.style={with q kv,
      label={[anchor=north east,xshift=-2pt,yshift=0pt]input q:\tiny Q},
      label={[anchor=north east,xshift=-2pt,yshift=0pt]input kv:\tiny KV},
    },
    flow/.style={very thick},
    tshape/.style={pos=0,anchor=south west,inner sep=2pt,xshift=1pt},
  ]

\path
node[val] (common S) at (0,0) {$S_{1:T-1}$}
node[next,op] (common embed) {Embeddings}
node[next,op,minimum height=1.35cm,text height=18pt] (common blocks) {Decoder Causal\\Transformer Block}
node[above=2.25cm of \tikzlastnode,enc,qkv,minimum height=1.35cm,text height=18pt] (encoder block) {Encoder Non-Causal\\Transformer Block}
node[next,enc] (encoder readout) {Encoder read-out FC}
node[next,enc,dash pattern=on 2.5pt off 1.5pt] (binary mapper) {Binary mapper}
node[next,val] (common Z) {$Z$}
node[next,op] (postsampling fc) {Post-sampler FC}
node[above=2.25cm of \tikzlastnode,op,qkv,minimum height=1.35cm,text height=18pt] (decoder block) {Decoder Causal\\Transformer Block}
node[next,op,minimum height=1.35cm,text height=18pt] (decoder blocks) {Decoder Causal\\Transformer Block}
node[next,op] (decoder readout) {Decoder read-out FC}
node[val,next] (decoder logits) {Logits $S_{2:T}$}
node[enc,minimum width=1.25cm,below=0.85cm of encoder block.input q] (zeta) {$\zeta$}
;

\draw[->,flow] (common S) -- (common embed)  node[tshape]{\scriptsize $T$} -- (common blocks) node[tshape]{\scriptsize $T \times D$};
\draw[->,flow] (decoder block) -- (decoder blocks) node[tshape]{\scriptsize $T \times D$};
\draw[->,flow] (decoder blocks) -- (decoder readout) node[tshape]{\scriptsize $T \times D$} -- (decoder logits) node[tshape]{\scriptsize $T \times V$};

\coordinate (after first half) at (common blocks|-zeta); 
\coordinate (right side) at ($(common Z.east)+(0.75, 0)$);

\draw[->,flow] (zeta) -- (encoder block.input q) node[tshape]{\scriptsize $T \times D$};
\coordinate (after encoder) at ($(binary mapper)+(0,1)$);
\coordinate (left side) at ($(common Z.west)+(-1,0)$);
\draw[->,flow] (encoder block) -- (encoder readout)  node[tshape]{\scriptsize $T \times D$} -- (binary mapper)  node[tshape]{\scriptsize $T \times H$} -- (common Z) node[tshape]{\scriptsize $T \times 2^H$} -- (postsampling fc) node[tshape]{\scriptsize $T \times 2^H$};

\coordinate (after sampling) at ($(postsampling fc.north)+(0, 0.75)$);
\node (sum) at ($(decoder block.input kv)+(0,-0.75)$) {$+$};

\draw[->,flow] (common blocks) -- (after first half) node[tshape]{\scriptsize $T \times D$} -| (right side) |- (after sampling) -| (decoder block.input q);
\draw[->,flow] (after first half-|encoder block.input kv) -- (encoder block.input kv);
\draw[->,flow] (after sampling-|sum) -- (sum) -- (decoder block.input kv);
\draw[->,flow,preaction={draw=white,line width=6pt,-,}] (postsampling fc) -- (after sampling)  node[tshape]{\scriptsize $T \times D$} |- (sum);

\coordinate (left info) at ($(decoder block.west)+(-3em,0)$);
\coordinate (right info) at ($(common blocks.east)+(4em,0)$);

\begin{pgfinterruptboundingbox}
\node[anchor=west,inner sep=1pt,xshift=1pt,yshift=0pt] at (common blocks.east) {$\times L/2$};
\node[anchor=west,inner sep=1pt,xshift=1pt,yshift=0pt] at (decoder blocks.east) {$\times L/2-1$};
\end{pgfinterruptboundingbox}


\end{tikzpicture}%


\caption{%
The Free Transformer. We omit the normalization layers and residual
connections from the model and the batch size from the tensor shapes
for clarity. The operators in orange are specific to the encoder and
are evaluated only for training or KV cache pre-filling, those with a
dashed contour have no trainable parameters. The Binary Mapper is
described in \S~\ref{sec:binary-mapper}. During generation, the
encoder is not evaluated and $Z$ is sampled uniformly among the
one-hot vectors of dimension $2^H$.}\label{fig:free-transformer}

\end{figure}
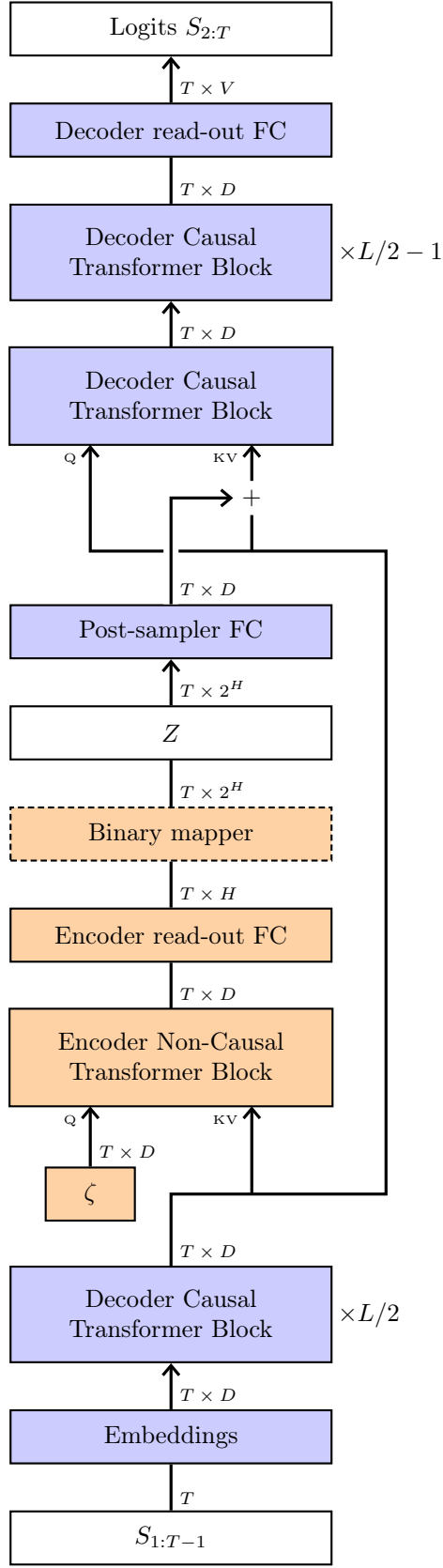

\subsection{Conditional Variational Autoencoder}\label{sec:cond-vari-auto}

Generating a full sequence from scratch with a model that depends on a
random variable $Z$ is trivial: sample $Z \sim P(Z)$ and then run the
standard auto-regressive process, with the computation of the logits
modulated by $Z$.

Training the model, however, is far more involved. Given a training
sample $S$, the objective is to maximize
\begin{equation}
P(S) = \int_z P(S \mid Z=z) P(Z=z) dz,
\end{equation}
which can be estimated only if we can get $Z$s consistent with $S$,
which amounts to a complex inference problem if we want $Z$ to capture
meaningful structural properties of the sequence.

Providing those $Z$s is the role of the encoder of a Variational
Autoencoder \citep{vae_2013}, whose main purpose is to sample from a
``good'' distribution $Q(Z \mid S)$ so that a sampled $Z$ modulates
the decoder in a way that leads it to generate $S$.

We follow this approach and optimize jointly the parameters of the
decoder and the parameters of a second model, which is an encoder in
the VAE sense.

Even though the noise $Z$ has no relation to $S$ initially, if the
training succeeds, the model will use it to structure the generative
process. In the example of a movie review generator of the previous
section, for instance, given a review from the training set, the
encoder would implicitly classify it as positive or negative, and
generate a consistent $Z$. Increasing $P(S \mid Z)$ with that $Z$
could be interpreted as improving the ``negative review generator'' or
the ``positive review generator'' that are implicitly encoded in the
decoder's weights.

A key element of this approach is to limit the amount of information
flowing from the encoder to the decoder through $Z$, so that the
encoder does not provide quantities that should be computed by the
decoder. At the limit the encoder could copy entirely $S$ into $Z$ so
that a trivial decoder, useless without the encoder, hence in
inference, would score perfectly in training.

The formal derivation of the VAE shows that the proper measure of
information is the Kullback-Leibler divergence between $Q(Z \mid S)$
and $P(Z)$, and that the loss to minimize should sum it with the
reconstruction loss, which here is the usual cross-entropy.

\subsection{Model structure}\label{sec:model-structure}

In what follows, we call ``Transformer Block'' the usual combination of
a Multi-Head Attention layer and a MLP-like tokenwise module, with
normalisation layers and residual connections.

As pictured on Figure~\ref{fig:models} and \ref{fig:free-transformer},
the Free Transformer is a standard decoder with a noise $Z$ injected
in its middle layer. This allows to share half of the Transformer
block with the encoder, cutting down drastically the computational
overhead by having a single Transformer block that has to be computed
specifically for the encoder. Hence, as we will see, this model
possesses all the components of a decoder Transformer and has an
additional non-causal block and two linear layers for the
encoder. While we did not investigate what is the best depth to inject
$Z$, doing it too early would reduce the encoder's capacity, and doing
it too late would reduce the decoder's capacity to process the latent
variables.

For clarity, we omit in what follows the batch size in the tensor
shapes.

As a standard decoder Transformer, the Free Transformer processes a
sequence of tokens by first encoding them with the embedding table
into a tensor $X_0$ of shape $T \times D$.

Then it evaluates sequentially the first $L/2$ Transformer blocks to
get $X_{L/2}$ of same shape, and at this point, it samples a sequence
of one-hot vectors $Z = (Z_1, \dots, Z_t) \in \{ 0, 1 \}^{T \times
  C}$. During generation, this is done by sampling, for each $Z_t$, an
index $c$ uniformly in $\{ 0, \dots, C-1\}$, and then encoding it as a
one-hot vector of dimension $C$. During training or KV cache
pre-filling, $Z$ has to be consistent with the tokens of $S$ already
fixed, and the sampling is done with the encoder instead, as described
in \S~\ref{sec:encoder-loss}.

This tensor $Z$ is processed by a linear layer to obtain a tensor $R$
of shape $T \times D$. Then, the $L/2+1$th Transformer block gets as
input for queries the tensor $X_{L/2}$ and as input for keys and
values the tensor $X_{L/2}+R$. The rest of the Transformer blocks are
evaluated in sequence to get $X_L$ which is processed by the read-out
linear layer to obtain the logit tensor $L$ of shape $T \times V$,
where $V$ is the vocabulary suze.

\begin{algorithm}
\caption{Forward pass of a standard decoder Transformer}\label{alg:gpt}
\begin{algorithmic}[1]
\Procedure{Forward}{$tokens$}
\State $x \gets \operatorname{embeddings}(tokens)$
\For{$n=1,\dots,B$}
\State $x \gets \operatorname{blocks}[n](in=x)$
\EndFor
\State $logits \gets \operatorname{linear\_readout}(\operatorname{RMS\_norm}(x))$
\State \textbf{return} $logits$
\EndProcedure
\end{algorithmic}
\end{algorithm}
%

%
%
\begin{algorithm}
\caption{Forward pass of a Free Transformer}\label{alg:ft}
\begin{algorithmic}[1]
\Procedure{Forward}{$tokens$}
\State $x \gets \operatorname{\operatorname{embeddings}}(tokens)$
\For{$n=1,\dots,B/2$}
\State $x \gets \operatorname{blocks}[n](in=x)$
\EndFor
\If{$train \text{\ or\ } prefill$}
\State $y \gets \operatorname{encoder\_block}(in\_q=zeta, in\_kv=x)$
\State $o \gets \operatorname{encoder\_linear\_readout}(\operatorname{RMS\_norm}(y))$
\State $z \gets \operatorname{binary\_mapper}(o)$
\Else
\State $z \gets \operatorname{one\_hot}(\operatorname{uniform\_sampler}())$
\EndIf
\State $r \gets \operatorname{linear\_post\_sampler}(z)$
\State $x \gets \operatorname{blocks}[B/2+1](in\_q=x, in\_kv=x+r)$
\For{$n=B/2+1,\dots,B$}
\State $x \gets \operatorname{blocks}[n](in=x)$
\EndFor
\State $logits \gets \operatorname{linear\_readout}(\operatorname{RMS\_norm}(x))$
\State \textbf{return} $logits$
\EndProcedure
\end{algorithmic}
\end{algorithm}

\subsection{Encoder and Loss}\label{sec:encoder-loss}

As stated in the previous section, during training or KV cache
pre-filling, the tensor $Z$ is sampled with the encoder.

The Free Transformer possesses one Transformer block specific to the
encoder, which is non-causal, making the encoder as a whole
non-causal. This is necessary since the conditioning by the decoder
may have long-range effects, requiring the full sequence to be taken
into account to get a proper conditional distribution of the latent.

This encoder-specific block gets as input for the queries a trained
token embedding $\zeta$ replicated to match the sequence length, and
for the keys and values the output of the first half of the decoder's
blocks. The motivation for using a learned constant input for the
queries instead of the standard representation of the input sequence
is to prevent the encoder from building a token-wise mapping and make it
instead capture global properties of the sequence that may be more
transferable across tasks and data-sets.

A linear readout computes from the encoder block's output a vector of
dimension $H=16$ for every token. These components are interpreted as
logits of individual bit, used to sample a value in $\{0, \dots,
2^H-1\}$ which is encoded into a one-hot vector of dimension $2^H =
65,536$, with gradient pass-through, as described in
\S~\ref{sec:binary-mapper}.


Hence, the random embedding $Z$ is a sequence of $T$ one-hot vectors
$Z_t$ of dimension $2^H$. The prior distribution used for generation
is uniform $P(Z_t=z) = 1/2^H$, and $Q(Z \mid S=s)$ is the distribution
corresponding to the sampling with the encoder described above. The KL
divergence is then equal to
\begin{equation}
\kl\Big(Q(Z_t \mid S_1, \dots, S_T) \, \Big\| \, P(Z_t)\Big) =  H \log 2 + \sum_{z=1}^{2^H} Q(Z=z \mid S) \log Q(Z=z \mid S).
\end{equation}

We control it by adding it to the loss, and prevent its collapse by
using a token-wise free bits method \citep{freebits_2016}. This means
that we sum the KL divergence of individual $Z_t$ that are above a
threshold $\kappa$ and ignore the others.

This leads us to use for training loss the sum of the standard
cross-entropy and the following quantity
\begin{equation}
\frac{1}{T} \sum_{t=1}^T \max\Big(0, \, \kl\Big(Q(Z_t \mid S_1, \dots, S_T) \, \Big\| \, P(Z_t)\Big) - \kappa\Big),
\end{equation}
where the threshold $\kappa$ is an hyperparameter.

\subsection{Binary Mapper}\label{sec:binary-mapper}

The last linear layer of the encoder computes for every index $t$ of the
sequence being processed a vector $L_t = (L_{t,1}, \dots, L_{t,H}) \in
\mathbb{R}^H$, whose components are interpreted as the logits of
individual bits of a binary encoding.

The Binary Mapper samples those bits $B_{t,1},\dots,B_{t,H}$
independentely with
\begin{equation}
P(B_{t,h} = 1) = \frac{1}{1 + e^{-L_{t,h}}},
\end{equation}
and outputs a one-hot vector $Y_t$ of dimension $2^H$ corresponding to
the resulting value:
\begin{equation}
Y_{t,d} = \left\{\begin{array}{cl}
1 & \text{ \ if \ } d = 1 + \sum_{h=1}^H 2^{h-1} B_{h,t}\\
0 & \text{ \ otherwise.}
\end{array}\right.
\end{equation}

During training, the computation also propagates the gradient of the
probabilities of the $2^H$ values. If ${U}(d) = ({U}_1(d), \dots,
{U}_H(d)) \in \{ 0, 1 \}^H$ is the binary encoding of $d$, and we
define $G_t$ as
\begin{align*}
G_{t,d} 
& = P(B_t = {U}(d-1)) \\
& = \exp \left( \sum_h \log P(B_{t,h} = {U}_h(d-1))\right) \\
& = \exp \left( \sum_h (1- {U}_h(d-1)) \log \left(1-\frac{1}{1 + e^{-L_{t,h}}}\right) +  {U}_h(d-1) \log \left(\frac{1}{1 + e^{-L_{t,h}}}\right)\right),
\end{align*}
then the Binary Mapper outputs
\begin{equation}
Y_{t,d} + G_{t,d} - \operatorname{detach}(G_{t,d}),
\end{equation}
where $\forall x, \operatorname{detach}(x)=x$ and
$J_{\operatorname{detach}}(x)=0$.

The motivation for using a binary encoding instead of having the
encoder output $2^H$ logits directly is to facilitate the gradient
pass-through thanks to the monotonicity of the sigmoid.


%
%


\section{Experiments}\label{sec:experiments}

We first test the qualitative behavior of the Free Transformer on a
synthetic task in \S~\ref{sec:synthetic-dataset}, then compare it on
multiple benchmarks to baselines with 1.5B and 8B parameters models
for various KL divergence thresholds in
\S~\ref{sec:exploratory-results}, and finally assess the performance
gain of a 8B parameter model trained on 1T tokens in
\S~\ref{sec:results-with-1t}.

\subsection{Synthetic Dataset}\label{sec:synthetic-dataset}

To confirm that the Free Transformer indeed utilizes $Z$ to condition
its generative process, we designed a synthetic dataset and trained a
small Free Transformer with different free-bits thresholds. Doing so
allows to observe what aspects of the modeling are packed by the encoder
in $Z$.

Each sequence in our synthetic training set is generated as follows:
\begin{itemize}

\item start with a string of 64 underscores ``\verb+_+'',

\item pick an upper case letter and a position in the sequence at
  random, and replace the underscores there with a ``target'' made of
  the selected letter repeated 8 times,

\item replace any character with an exclamation mark with probability
  $1/16$

\item concatenate a prompt made of the target's letter followed by a
  ``\verb+>+''.

\end{itemize}
A few sequences generated with that process are shown in
Figure~\ref{fig:toy-examples}.

We trained a Free Transformer on this data for four different values
of the free bits threshold $\kappa$, and generated with the same
random prompt three groups of sequences with each model, as pictured
in Figure~\ref{fig:toy-results}. For each model, in the blue group, the noise
$Z$ is sampled independently for each sequence, whereas we sampled one
$Z$ only for each of the green groups, used to generate all its
sequences.

For very low values of the KL divergence, the model behaves like a
vanilla model (Figure~\ref{fig:toy-results}, middle left), and when the
value increases, the model encodes initially the position of the
target alone in the latent state (Figure~\ref{fig:toy-results}, middle right),
then encodes both the target position and the noise
(Figure~\ref{fig:toy-results}, bottom left), and finally encodes the full
sequence, resulting in incorrect generation (Figure~\ref{fig:toy-results},
bottom right).

\begin{figure} 
\center

\begin{SaveVerbatim}{myverb}
K>!_________!_______!_______!____________!_______KKKKKKKK_________
C>___CCCCCCCC_________________!______!_____________!__!__!________
X>___________________!!_________!!___XX!XXXXX_____!_______________
R>!__RRRRRRRR_____________!__!_______________________________!____
P>!__!___________________________________________________PPPPPPPP_
L>_______!_!LLLLLLLL________!___________!!____________!___________
V>__!_________________!__!________VVVVVV!V________!_____!____!____
P>_________PPPPPPPP_____!________________!_______________________!
A>_______!___________!___________________________!_______AAAAAAA!_
P>____________________!____PPPPPP!P____!___________!_________!__!!
I>__________________________________________!_!__IIIIIIII_________
D>______!_!___________________________!_________!DDDDDDD__________
A>_____!___AAAAAAA!_______________!_________________!______!______
J>_______!_____!_________J!JJJJJJ_____________!___________________
\end{SaveVerbatim}
\BUseVerbatim[fontsize=\notsotiny,baselinestretch=1.0,fontfamily=courier,fontseries=b]{myverb}

\vspace*{2ex}

%
%
\caption{The synthetic sequences of \S~\ref{sec:synthetic-dataset} are
  of fixed length, with a ``target'' made of a random letter repeated
  8 times at a random position, an i.i.d. noise of exclamation marks,
  and a prompt indicating the target's
  letter.}\label{fig:toy-examples}


\vspace*{6ex}

\begin{minipage}{0.49\textwidth}
\begin{center}



\begin{SaveVerbatim}{myverb}
T>_________________________________TTTTTTTT_______________________
T>________________________________TTTTTTTT________________________
T>_________________________________TTTTTTTT_______________________
T>_____________________________________!TTTTTTTT__________________
T>_____________________________________________!_______TTTTTTTT___
\end{SaveVerbatim}
\colorbox{lightblue}{\BUseVerbatim[fontsize=\notsotiny,baselinestretch=1.0,fontfamily=courier,fontseries=b]{myverb}}

\vspace*{1ex}

\begin{SaveVerbatim}{myverb}
T>_________________________________TTTTTTTT_______________________
T>_____________________________________________________TTTTTTTT___
T>______________________________________________________!TTTTTTTT_
T>___________________________!____________________________TTTTTTTT
T>_______________________________________________________TTTTTTTT_
\end{SaveVerbatim}
\colorbox{lightgreen}{\BUseVerbatim[fontsize=\notsotiny,baselinestretch=1.0,fontfamily=courier,fontseries=b]{myverb}}

\vspace*{1ex}

\begin{SaveVerbatim}{myverb}
T>_____________________________________________________TTTTTTTT___
T>_____________________________________________________TTTTTTTT___
T>___________________________________TTTTTTTT______!_____!____!___
T>_____________________________________________TTTTTTTT___________
T>__________________________________TTTTTTTT______________________
\end{SaveVerbatim}
\colorbox{lightgreen}{\BUseVerbatim[fontsize=\notsotiny,baselinestretch=1.0,fontfamily=courier,fontseries=b]{myverb}}

\vspace*{2ex}

$\kappa = \log(2)/64$ (1/64 bit)

\end{center}
\end{minipage}
\hspace*{\stretch{1}}
\begin{minipage}{0.49\textwidth}
\begin{center}



\begin{SaveVerbatim}{myverb}
F>_______________________________________________________FFFFFFFF_
F>___________________FFFFFFFF__________!__________!_______________
F>_________________FFFFFFFF________________________!____________!_
F>___________________________________FFFFFFFF__________________!__
F>____________________________________________!FFF!FFFF___________
\end{SaveVerbatim}
\colorbox{lightblue}{\BUseVerbatim[fontsize=\notsotiny,baselinestretch=1.0,fontfamily=courier,fontseries=b]{myverb}}

\vspace*{1ex}

\begin{SaveVerbatim}{myverb}
F>_______________________!____________________________FFFFFFFF____
F>____________________________________________________FFFFFFFF____
F>____________________________________________________FFFFFFFF!___
F>___________________________________________________FFFFFFFF!____
F>_____________________________________________________FFFFFFFF___
\end{SaveVerbatim}
\colorbox{lightgreen}{\BUseVerbatim[fontsize=\notsotiny,baselinestretch=1.0,fontfamily=courier,fontseries=b]{myverb}}

\vspace*{1ex}

\begin{SaveVerbatim}{myverb}
F>_________________________FFFFFFFF!_________________!____________
F>__________!____________FF!FFFFF_________________________________
F>________________________FFFFFFFF___________________!____________
F>_______________________FFFFFFFF_______________!______!__________
F>_______________________FFFFFFFF______________________!__________
\end{SaveVerbatim}
\colorbox{lightgreen}{\BUseVerbatim[fontsize=\notsotiny,baselinestretch=1.0,fontfamily=courier,fontseries=b]{myverb}}

\vspace*{2ex}

$\kappa = \log(2)/8$ (1/8 bit)

\end{center}
\end{minipage}

\vspace*{6ex}


\begin{minipage}{0.49\textwidth}
\begin{center}

\begin{SaveVerbatim}{myverb}
J>JJJJJJJ!____!_________!!__!_!_!__________!___________!___!__!___
J>_____!_____!______!______!_JJJJJJJJ______________________!______
J>___JJJ!JJJJ____________!__!___!_!__!_____!_____!!__!_____!___!__
J>__!___________JJJJJJJJ___________________!________!____!________
J>______!!___!!!_____________JJJJJJJJ!______!!!_!_!___!___________
\end{SaveVerbatim}
\colorbox{lightblue}{\BUseVerbatim[fontsize=\notsotiny,baselinestretch=1.0,fontfamily=courier,fontseries=b]{myverb}}

\vspace*{1ex}

\begin{SaveVerbatim}{myverb}
J>_________JJJ!JJJJ__!______________!__________!!___!!_________!__
J>_________JJJ!JJJJ__!______________!______!____!__!!__________!__
J>_________JJ!JJJJJ__!_______!________!________!!__!!__________!__
J>_________JJJ!JJJJ__!________________!!____!!_!!____!_________!__
J>___!_____JJ!JJJJJ__!______________!_______!__!!_______!______!__
\end{SaveVerbatim}
\colorbox{lightgreen}{\BUseVerbatim[fontsize=\notsotiny,baselinestretch=1.0,fontfamily=courier,fontseries=b]{myverb}}

\vspace*{1ex}

\begin{SaveVerbatim}{myverb}
J>__JJJJJJJJ______!___________!____!_!______!_______!__!________!_
J>__JJJJJJJJ______!___________!____!_!________!__!__!__!________!_
J>_JJJJJJJJ_______!_______!________!_!______!_______!__!_______!__
J>_JJJJJJJJ_______!________________!_!________!__!____!!___!____!_
J>_JJJJJJJJ_______!___________!_____!_!_____!_______!__!_______!__
\end{SaveVerbatim}
\colorbox{lightgreen}{\BUseVerbatim[fontsize=\notsotiny,baselinestretch=1.0,fontfamily=courier,fontseries=b]{myverb}}

\vspace*{2ex}

$\kappa = \log(2)$ (1 bit)

\end{center}
\end{minipage}
\hspace*{\stretch{1}}
\begin{minipage}{0.49\textwidth}
\begin{center}

\begin{SaveVerbatim}{myverb}
O>___________!!________!__!________________!_____!_______!______!!
O>____OOOOO______________________________________________________!
O>O!___O_!__!_!__!__!_OO____!!__OO_________________!!________!___!
O>_____!_____!_____________!_______________!_F___!!_!_______!_____
O>__OOOO!OO!___OO____!____O_!________________________O!____O_____!
\end{SaveVerbatim}
\colorbox{lightblue}{\BUseVerbatim[fontsize=\notsotiny,baselinestretch=1.0,fontfamily=courier,fontseries=b]{myverb}}

\vspace*{1ex}

\begin{SaveVerbatim}{myverb}
O>_________OOOO________O__________!_____________________!!_!O____!
O>_________OO!O________O____O_____!_____________________!!!!O____!
O>_________OO!O________O____O_____!________________O____!__!O____!
O>_________OOOO________O____O_____!_____________________!!_!O____!
O>_________OOOO________O____O_____!______________________!_!O____!
\end{SaveVerbatim}
\colorbox{lightgreen}{\BUseVerbatim[fontsize=\notsotiny,baselinestretch=1.0,fontfamily=courier,fontseries=b]{myverb}}

\vspace*{1ex}

\begin{SaveVerbatim}{myverb}
O>__!___OO______________!___OO!O________O!O____________O_____O___!
O>O_!__OOO__________________OOOO________O!O____________O________!!
O>__!___OO__________________OO__________O!O____________O_____O____
O>__!___OO__________________OO__________O!O____O_______O_____O__!!
O>O_!__OOO_________!OO__!___OO__________O!O____________O_____O___!
\end{SaveVerbatim}
\colorbox{lightgreen}{\BUseVerbatim[fontsize=\notsotiny,baselinestretch=1.0,fontfamily=courier,fontseries=b]{myverb}}

\vspace*{2ex}

$\kappa = 8 \log(2)$ (8 bits)

\end{center}
\end{minipage}

\vspace*{2ex}

%
%
\caption{Results with a Free Transformer trained on the synthetic
  sequences of \S~\ref{sec:synthetic-dataset} for different prompts
  and free bit thresholds. To investigate the information encoded in
  the latent tensor, we sample a $Z$ per sequence of a blue box, and a
  $Z$ per green box. For very low values of the KL divergence, the
  model behaves like a vanilla model (top left), and when the KL
  divergence increases, the model encodes initially the position of
  the target alone in the latent state (top right), then encodes both
  the target position and the noise (bottom left), and finally encodes
  the full sequence, resulting in incorrect generation (bottom
  right).}\label{fig:toy-results}

\end{figure}

\subsection{Baseline architectures}

For assessing performance on standard benchmarks we used decoder-only
Transformers implemented in the same Meta FAIR Transformer codebase as
the one used by \cite{cwm2025} for the Computational World
Model. Those are well optimized models using the SwiGLU non-linearity
\citep{shazeer2020glu}, pre-normalization with RMSNorm
\citep{zhang2019rmsnorm}, Rotary Positional Embedding (RoPE,
\citealt{su2021rope}), and Group Query Attention (GQA,
\citealt{ainslie2023gqa}). The vocabulary size is $2^{17} \approx
130k$.

We used two sizes of models:
\begin{itemize}

\item A 1.5B model, with $28$ layers, weight tying between the
  embeddings and the logit readout, model dimension $1536$, $12$ query
  heads, and $2$ key-value heads. It is trained with 47B tokens, which
  requires 32 H100s for $\approx$ 12 hours.

\item A 8B model with the structure of a Llama-3, which is $32$
  layers, model dimension $4096$, $32$ query heads, and $8$ key-value
  heads. It is trained with 200B tokens which requires 256 H100s for
  $\approx$ 24 hours, or with 1T tokens, which takes 5 days.

\end{itemize}

We compare those baselines to the equivalent Free Transformers, which
require one additional layer for the encoder during training and KV
cache pre-filling, resulting in a compute and memory overhead of $1/28
\approx 3.6\%$ for the 1.5B and $1/32 \approx 3.1\%$ for the 8B.

\subsection{Setup and hyperparameters}

We kept our findings as clear as possible by avoiding other sources of
performance improvement:
\begin{itemize}

\item We stuck to the baseline architecture, optimizer, and learning
  rate schedule that were used to train the baselines in FAIR's
  framework, and did not optimize any hyperparameter for our setup.

\item We avoided any recipes for the VAE components, such as removing
  sampling in inference. We followed the formal expressions rigorously.

\item We fixed $H$ to $16$ so that the dimention of $Z_t$ was
  comparable to the vocabulary size of $2^{17}$.

\end{itemize}


We stress that the optimization hyperparameters were highly tuned for
the baselines, and it is probable that a combination of an encoder and
a decoder has specific requirements that would greatly benefit from an
adapted training procedure.


\begin{table}


\center
\footnotesize
\setlength{\tabcolsep}{2pt}
\renewcommand{\arraystretch}{1.2}

\begin{tabular}{l|c|cc|cc|cc|cc}
\multicolumn{10}{c}{\textbf{1.5B models (47B tokens)}}\\
\hline
& \multirow{2}{*}{\textbf{Baseline}} & \multicolumn{8}{c}{\textbf{\modelname}} \\
& & \multicolumn{2}{c|}{1/4 bit}&\multicolumn{2}{c|}{1/2 bit}& \multicolumn{2}{c|}{1 bit} & \multicolumn{2}{c}{2 bits} \\
%
%
\hline
\multicolumn{10}{c}{Generative code/math}\\
\hline
human\_eval\_plu (pass@1) & 0.055 & 0.079 & \colorbox{green}{+44.44\%} & 0.079 & \colorbox{green}{+44.44\%} & 0.085 & \colorbox{green}{+55.56\%} & 0.085 & \colorbox{green}{+55.56\%}\\
mbpp (pass@1) & 0.112 & 0.144 & \colorbox{green}{+28.57\%} & 0.148 & \colorbox{green}{+32.14\%} & 0.152 & \colorbox{green}{+35.71\%} & 0.122 & \colorbox{green!25}{+8.93\%}\\
gsm8k (em) & 0.025 & 0.028 & \colorbox{green}{+12.12\%} & 0.027 & \colorbox{green!25}{+6.06\%} & 0.033 & \colorbox{green}{+30.30\%} & 0.027 & \colorbox{green!25}{+6.06\%}\\
\hline
\multicolumn{10}{c}{Multi-choice general knowledge / common sense}\\
\hline
mmlu (macro\_avg/acc\_char) & 0.252 & 0.265 & \colorbox{green!25}{+5.31\%} & 0.261 & \colorbox{green!25}{+3.76\%} & 0.254 & \colorbox{green!25}{+1.07\%} & 0.257 & \colorbox{green!25}{+2.19\%}\\
csqa (acc\_char) & 0.199 & 0.175 & \colorbox{red!80}{-11.93\%} & 0.199 & \colorbox{white}{+0.00\%} & 0.187 & \colorbox{red!10}{-6.17\%} & 0.197 & \colorbox{white}{-0.82\%}\\
hellaswag (acc\_char) & 0.593 & 0.591 & \colorbox{white}{-0.40\%} & 0.594 & \colorbox{white}{+0.15\%} & 0.592 & \colorbox{white}{-0.27\%} & 0.595 & \colorbox{white}{+0.32\%}\\
winogrande (acc\_char) & 0.603 & 0.604 & \colorbox{white}{+0.13\%} & 0.598 & \colorbox{white}{-0.79\%} & 0.600 & \colorbox{white}{-0.52\%} & 0.597 & \colorbox{red!10}{-1.05\%}\\
obqa (acc\_completion) & 0.446 & 0.450 & \colorbox{white}{+0.90\%} & 0.468 & \colorbox{green!25}{+4.93\%} & 0.460 & \colorbox{green!25}{+3.14\%} & 0.490 & \colorbox{green!25}{+9.87\%}\\
arc\_challenge (acc\_completion) & 0.400 & 0.392 & \colorbox{red!10}{-1.93\%} & 0.386 & \colorbox{red!10}{-3.43\%} & 0.405 & \colorbox{green!25}{+1.29\%} & 0.385 & \colorbox{red!10}{-3.65\%}\\
arc\_easy (acc\_completion) & 0.596 & 0.602 & \colorbox{white}{+0.92\%} & 0.592 & \colorbox{white}{-0.64\%} & 0.603 & \colorbox{green!25}{+1.06\%} & 0.592 & \colorbox{white}{-0.71\%}\\
piqa (acc\_char) & 0.734 & 0.736 & \colorbox{white}{+0.22\%} & 0.738 & \colorbox{white}{+0.52\%} & 0.734 & \colorbox{white}{+0.07\%} & 0.733 & \colorbox{white}{-0.15\%}\\
\hline
\multicolumn{10}{c}{Multi-choice text understanding}\\
\hline
race.high (acc\_char) & 0.390 & 0.382 & \colorbox{red!10}{-2.20\%} & 0.390 & \colorbox{white}{+0.00\%} & 0.387 & \colorbox{white}{-0.81\%} & 0.386 & \colorbox{red!10}{-1.03\%}\\
race.middle (acc\_char) & 0.532 & 0.511 & \colorbox{red!10}{-3.93\%} & 0.519 & \colorbox{red!10}{-2.49\%} & 0.522 & \colorbox{red!10}{-1.83\%} & 0.514 & \colorbox{red!10}{-3.40\%}\\
boolq (acc\_completion) & 0.583 & 0.632 & \colorbox{green!25}{+8.39\%} & 0.614 & \colorbox{green!25}{+5.35\%} & 0.648 & \colorbox{green}{+11.12\%} & 0.620 & \colorbox{green!25}{+6.29\%}\\
\hline
\multicolumn{10}{c}{Culture}\\
\hline
nq (em) & 0.081 & 0.069 & \colorbox{red!80}{-15.36\%} & 0.073 & \colorbox{red!10}{-9.56\%} & 0.075 & \colorbox{red!10}{-7.17\%} & 0.071 & \colorbox{red!80}{-11.95\%}\\
tqa (em) & 0.205 & 0.191 & \colorbox{red!10}{-6.93\%} & 0.190 & \colorbox{red!10}{-7.58\%} & 0.200 & \colorbox{red!10}{-2.84\%} & 0.197 & \colorbox{red!10}{-4.13\%}
 \\
\hline
\end{tabular}

\bigskip

%
%
\caption{Performance of 1.5B models trained on 47B tokens. The
  training procedure was tuned for the baseline and kept unchanged,
  but the Free Transformers require $3.6\%$ more compute and
  parameters for the encoder. See Figure~\ref{fig:smodel-perfs} in
  Appendix \ref{sec:perf-during-train} for the performance during
  training.}\label{tab:smodel-perfs}





\bigskip

\begin{tabular}{l|c|cc|cc|cc|cc}
\multicolumn{10}{c}{\textbf{8B models (200B tokens)}}\\
\hline
& \multirow{2}{*}{\textbf{Baseline}} & \multicolumn{8}{c}{\textbf{\modelname}} \\
& & \multicolumn{2}{c|}{1/4 bit}& \multicolumn{2}{c|}{1/2 bit} & \multicolumn{2}{c}{1 bit} & \multicolumn{2}{c}{2 bits} \\
%
\hline
\multicolumn{10}{c}{Generative code/math}\\
\hline
human\_eval\_plu (pass@1) & 0.159 & 0.171 & \colorbox{green!25}{+7.69\%} & 0.189 & \colorbox{green}{+19.23\%} & 0.165 & \colorbox{green!25}{+3.85\%} & 0.177 & \colorbox{green}{+11.54\%}\\
mbpp (pass@1) & 0.278 & 0.330 & \colorbox{green}{+18.71\%} & 0.306 & \colorbox{green}{+10.07\%} & 0.298 & \colorbox{green!25}{+7.19\%} & 0.318 & \colorbox{green}{+14.39\%}\\
gsm8k (em) & 0.086 & 0.079 & \colorbox{red!10}{-8.77\%} & 0.095 & \colorbox{green!25}{+9.65\%} & 0.104 & \colorbox{green}{+20.18\%} & 0.096 & \colorbox{green}{+10.53\%}\\
\hline
\multicolumn{10}{c}{Multi-choice general knowledge / common sense}\\
\hline
mmlu (macro\_avg/acc\_char) & 0.359 & 0.337 & \colorbox{red!10}{-6.13\%} & 0.398 & \colorbox{green}{+10.97\%} & 0.365 & \colorbox{green!25}{+1.81\%} & 0.345 & \colorbox{red!10}{-4.00\%}\\
csqa (acc\_char) & 0.356 & 0.292 & \colorbox{red!80}{-17.93\%} & 0.450 & \colorbox{green}{+26.21\%} & 0.346 & \colorbox{red!10}{-2.99\%} & 0.324 & \colorbox{red!10}{-8.97\%}\\
hellaswag (acc\_char) & 0.735 & 0.737 & \colorbox{white}{+0.26\%} & 0.737 & \colorbox{white}{+0.26\%} & 0.732 & \colorbox{white}{-0.45\%} & 0.738 & \colorbox{white}{+0.39\%}\\
winogrande (acc\_char) & 0.680 & 0.667 & \colorbox{red!10}{-1.86\%} & 0.664 & \colorbox{red!10}{-2.32\%} & 0.664 & \colorbox{red!10}{-2.32\%} & 0.667 & \colorbox{red!10}{-1.86\%}\\
obqa (acc\_completion) & 0.522 & 0.508 & \colorbox{red!10}{-2.68\%} & 0.484 & \colorbox{red!10}{-7.28\%} & 0.530 & \colorbox{green!25}{+1.53\%} & 0.554 & \colorbox{green!25}{+6.13\%}\\
arc\_challenge (acc\_completion) & 0.465 & 0.483 & \colorbox{green!25}{+3.87\%} & 0.468 & \colorbox{white}{+0.55\%} & 0.452 & \colorbox{red!10}{-2.95\%} & 0.485 & \colorbox{green!25}{+4.24\%}\\
arc\_easy (acc\_completion) & 0.677 & 0.676 & \colorbox{white}{-0.25\%} & 0.665 & \colorbox{red!10}{-1.81\%} & 0.668 & \colorbox{red!10}{-1.44\%} & 0.679 & \colorbox{white}{+0.31\%}\\
piqa (acc\_char) & 0.774 & 0.780 & \colorbox{white}{+0.77\%} & 0.782 & \colorbox{green!25}{+1.05\%} & 0.785 & \colorbox{green!25}{+1.41\%} & 0.793 & \colorbox{green!25}{+2.46\%}\\
\hline
\multicolumn{10}{c}{Multi-choice text understanding}\\
\hline
race.high (acc\_char) & 0.433 & 0.447 & \colorbox{green!25}{+3.30\%} & 0.443 & \colorbox{green!25}{+2.25\%} & 0.444 & \colorbox{green!25}{+2.58\%} & 0.435 & \colorbox{white}{+0.53\%}\\
race.middle (acc\_char) & 0.594 & 0.592 & \colorbox{white}{-0.35\%} & 0.591 & \colorbox{white}{-0.47\%} & 0.587 & \colorbox{red!10}{-1.17\%} & 0.584 & \colorbox{red!10}{-1.64\%}\\
boolq (acc\_completion) & 0.705 & 0.632 & \colorbox{red!80}{-10.37\%} & 0.632 & \colorbox{red!80}{-10.33\%} & 0.687 & \colorbox{red!10}{-2.47\%} & 0.671 & \colorbox{red!10}{-4.82\%}\\
\hline
\multicolumn{10}{c}{Culture}\\
\hline
nq (em) & 0.181 & 0.183 & \colorbox{green!25}{+1.38\%} & 0.167 & \colorbox{red!10}{-7.67\%} & 0.173 & \colorbox{red!10}{-4.14\%} & 0.168 & \colorbox{red!10}{-6.90\%}\\
tqa (em) & 0.440 & 0.438 & \colorbox{white}{-0.28\%} & 0.443 & \colorbox{white}{+0.80\%} & 0.434 & \colorbox{red!10}{-1.19\%} & 0.446 & \colorbox{green!25}{+1.45\%}
 \\
\hline
\end{tabular}

%
%
\caption{Performance of 8B models trained on 200B tokens. The training
  procedure was tuned for the baseline and kept unchanged, but the
  Free Transformers require $3.1\%$ more compute and parameters for
  the encoder. See Figure~\ref{fig:llama-perfs} in Appendix
  \ref{sec:perf-during-train} for the performance during
  training.}\label{tab:llama-perfs}

\end{table}

\begin{table}


\center
\footnotesize
\setlength{\tabcolsep}{2pt}
\renewcommand{\arraystretch}{1.2}

\begin{tabular}{l|c|cc|c|cc}
\multicolumn{7}{c}{\textbf{8B models (1T tokens)}}                                                                                                                                 \\
\hline
 & \multicolumn{3}{c|}{\textbf{Final value}} & \multicolumn{3}{c}{\textbf{Average (last third)}}                                                                                   \\
\hline
 & \multirow{2}{*}{\textbf{Baseline}}        & \multicolumn{2}{c|}{\textbf{Free Transformer}} & \multirow{2}{*}{\textbf{Baseline}} & \multicolumn{2}{c}{\textbf{Free Transformer}} \\
 &                                           & \multicolumn{2}{c|}{1/2 bit}                   &                                    & \multicolumn{2}{c}{1/2 bit}                   \\
%
%
\hline
\multicolumn{7}{c}{Generative code/math}\\
\hline
human\_eval\_plu (pass@1) & 0.268 & 0.299 & \colorbox{green}{+11.36\%}& 0.245 & 0.256 & \colorbox{green!25}{+4.22\%}\\
mbpp (pass@1) & 0.428 & 0.440 & \colorbox{green!25}{+2.80\%}& 0.396 & 0.421 & \colorbox{green!25}{+6.08\%}\\
gsm8k (em) & 0.321 & 0.331 & \colorbox{green!25}{+2.83\%}& 0.280 & 0.296 & \colorbox{green!25}{+5.84\%}\\
\hline
\multicolumn{7}{c}{Multi-choice general knowledge / common sense}\\
\hline
mmlu (macro\_avg/acc\_char) & 0.592 & 0.623 & \colorbox{green!25}{+5.20\%}& 0.567 & 0.596 & \colorbox{green!25}{+5.16\%}\\
csqa (acc\_char) & 0.707 & 0.748 & \colorbox{green!25}{+5.79\%}& 0.689 & 0.733 & \colorbox{green!25}{+6.28\%}\\
hellaswag (acc\_char) & 0.799 & 0.799 & \colorbox{white}{-0.01\%}& 0.787 & 0.788 & \colorbox{white}{+0.18\%}\\
winogrande (acc\_char) & 0.739 & 0.735 & \colorbox{white}{-0.53\%}& 0.725 & 0.727 & \colorbox{white}{+0.27\%}\\
obqa (acc\_completion) & 0.564 & 0.562 & \colorbox{white}{-0.35\%}& 0.556 & 0.551 & \colorbox{white}{-0.86\%}\\
arc\_challenge (acc\_completion) & 0.542 & 0.535 & \colorbox{red!10}{-1.42\%}& 0.524 & 0.522 & \colorbox{white}{-0.40\%}\\
arc\_easy (acc\_completion) & 0.721 & 0.711 & \colorbox{red!10}{-1.41\%}& 0.706 & 0.711 & \colorbox{white}{+0.68\%}\\
piqa (acc\_char) & 0.805 & 0.812 & \colorbox{white}{+0.88\%}& 0.802 & 0.807 & \colorbox{white}{+0.61\%}\\
\hline
\multicolumn{7}{c}{Multi-choice text understanding}\\
\hline
race.high (acc\_char) & 0.473 & 0.463 & \colorbox{red!10}{-2.06\%}& 0.467 & 0.460 & \colorbox{red!10}{-1.55\%}\\
race.middle (acc\_char) & 0.632 & 0.634 & \colorbox{white}{+0.33\%}& 0.623 & 0.624 & \colorbox{white}{+0.16\%}\\
boolq (acc\_completion) & 0.713 & 0.725 & \colorbox{green!25}{+1.63\%}& 0.755 & 0.754 & \colorbox{white}{-0.10\%}\\
\hline
\multicolumn{7}{c}{Culture}\\
\hline
nq (em) & 0.248 & 0.247 & \colorbox{white}{-0.22\%}& 0.229 & 0.227 & \colorbox{white}{-0.76\%}\\
tqa (em) & 0.583 & 0.577 & \colorbox{red!10}{-1.00\%} & 0.549 & 0.544 & \colorbox{white}{-0.90\%}
                                                                                                                                        \\
\hline
\end{tabular}

%
%
\caption{Performance of 8B models trained on 1T tokens. We also
  provide the average over the last third of the iterations to
  mitigate the irregularity of the performance increase during
  training and get a more accurate estimate of the relative
  improvement. The optimization hyperparameters were tuned for the
  baseline and kept unchanged, but the Free Transformers require
  $3.1\%$ more compute and parameters for the encoder. See
  Figure~\ref{fig:llama-perfs-1t} in Appendix
  \ref{sec:perf-during-train} for the performance during
  training.}\label{tab:llama-perfs-1t}

\end{table}


\subsection{Exploratory Results}\label{sec:exploratory-results}

We ran a series of experiments to assess the general behavior of the
Free Transformer, and to calibrate the $\kappa$ threshold.

For any value of $\kappa$, the cross-entropy goes down regularly
during training, with no more instability and spikes than what happens
with the baselines. The KL divergence rapidly goes under $\kappa$ and
stays there. When we compare the cross-entropies for various $\kappa$,
they go down when $\kappa$ increases as expected, but the values
remain extremely close, with a difference of the order of $0.01$ for a
cross-entropy of $\approx 2$ for the 1.5B and $\approx 1.8$ for the
8B.

For both sizes of models, setting $\kappa = 4 \log 2$, corresponding
to $4$ bits of information per token, resulted in a collapse of the
cross-entropy, indicating that the encoder found a way to channel
fully the tokens to predict, and resulting in a collapse of
performance on the downstream tasks. It is noteworthy that the
baseline 8B model reaches during training a cross-entropy of $1.8 =
2.59 \log(2)$, hence may explain why allowing $2$ bits does not
collapse, while allowing $4$ bits does.

The performance on downstream tasks are given in Table
\ref{tab:smodel-perfs} for the 1.5B models, and Table
\ref{tab:llama-perfs} for the 8B models, both for four different
values of $\kappa$ corresponding to $1/2$ to $2$ bits of information
per token. Graphs of performance during training are given in Appendix
\ref{sec:perf-during-train} in Figures \ref{fig:smodel-perfs} and
\ref{fig:llama-perfs}.


We observe a substantial increase of performance on HumanEval+, MBPP,
and GSM8K which are arguably the benchmarks requiring some form of
reasoning, and there also is a clear improvement for the 8B model with
1/2 bit of KL divergence on MMLU and CSQA, which are multi-choice
questions.


\subsection{Results with 1T tokens training}\label{sec:results-with-1t}

To measure improvement in a more realistic setting, closer to models
actually used in real applications, we trained 8B models on 1T tokens,
which improves drastically the performance of both the baseline and
the Free Transformer.

Given the results with 200B tokens, we chose the value $\kappa =
\log(2)/2$ corresponding to half a bit of information per token at
most.

The performance on downstream tasks are given in Table
\ref{tab:llama-perfs-1t} and the corresponding graphs during training
in Figure \ref{fig:llama-perfs-1t} of Appendix
\ref{sec:perf-during-train}. We provide in the table the performance
measured at the end of the training as for the other configurations,
but in addition we also give the average over the last third of the
training. We can observe on the graphs that the rate of improvement
tend to be constant on this interval, which justifies averaging to
mitigate the performance fluctuations.

The key result is the boost of performance on HumanEval+, MBPP, GSM8K,
MMLU and CSQA, confirming what we observed in the smaller settings,
and a greater stability on other tasks.


\section{Previous work}

There have been several attempts at combining a VAE and a decoder
Transformer, generally with a focus on improving topic models and
providing ways to guide the generation.

The OPTIMUS model \citep{li2020optimus} combines a pre-trained BERT as
text embedding / encoder, with a GPT-2 playing the role of decoder,
which are fine-tuned with a VAE-like loss.

The latent embedding $Z$ is computed thanks to a CLS token, that is by
adding a token to the input and a read-out to extract its embedding in
the output. To modulate the GPT-2 generation with it, it is either (1)
concatenated as an additional token in every layer, or (2) added to
the input token embeddings. Collapse of the KL divergence is prevented
during training with the free bits method \citep{freebits_2016}.

This approach allows for better guided text generation with GPT-2 and
better generalization on low-data languages with BERT.

\citet{xie2021} extend OPTIMUS with a multi-objective loss, adding in
particular the prediction of the story topic, using the output of
another model as ground truth, to obtain a better embedding space.

The CVAE proposed by \citet{fang2021cvae} combines two pre-trained
GPT-2, one used as the encoder without causal masking. The embedding $Z$
is an average of the encoder's output, and the authors propose three
ways to modulate the decoder with linear images of it: (1) add it to
each input token embedding, (2) concatenate it to the Ks and Vs in
every layer, (3) add it before the softmax. Experiments demonstrate
that this method allows controlling the generation without hurting the
quality of the result.

AdaVAE \citep{tu2022adavae} is similarly the combination of two
pre-trained GPT-2, the first without causal masking playing the role
of the encoder. The latent embedding $Z$ is extracted from its output
with a slightly modified attention operator. It is then injected into
the decoder by either concatenating an image of it to the keys and
values as in OPTIMUS, or before the softmax as in CVAE.




\section{Conclusion}

The Free Transformer is a direct extension of a standard decoder
Transformer, with the abstract structure of a conditional VAE. It is
implemented with a single additional non-causal Transformer block and
requires a few percent of computational and memory usage overhead.

Its structure makes it able to learn latent random variables
unsupervised, and to condition its generative process on them. In some
ways, this approach aims at achieving in latent space with an
autoencoder what reasoning models do with chains-of-thought in token
space and an RL procedure \citep{deepseer12025}. A combination of the
two is, of course, promising.

The performance boost without tuning the optimization hyperparameters
across multiple benchmarks and two sizes of models, is a strong signal
that the overall approach actually improves the inductive bias of the
vanilla Transformer.

Many properties and design choices should be explored. The performance
curves during training are often unstable, possibly due to the
coupling of the optimization of the encoder and the decoder, and using
different optimization methods could be fruitful. The random embedding
itself could take many forms, and the one used in our implementation
is arbitrary.

Finally, the behavior in larger scales, both in parameter count and
dataset size, remains to be investigated.

%

\bibliographystyle{plainnatmodified}
\newcommand{\biburl}[1]{\href{#1}{pdf}}

\bibliography{free-transformer}


\appendix

%
%

\section{Evaluation Benchmarks}

\begin{itemize}

\item {HellaSwag}: Multiple choices. Common sense focusing on physically situated scenarios. \citep{zellers2019hellaswag}

\item {WinoGrande}: Large-scale adversarial Winograd-style pronoun resolution (fill-in-the-blank) designed to reduce annotation artifacts. \citep{sakaguchi2019winograndeadversarialwinogradschema}

\item {ARC (AI2 Reasoning Challenge)}: Grade-school science multiple choice. \citep{clark2018thinkarc}

\item {PIQA}: Physical commonsense multiple choice about everyday goals and affordances. \citep{bisk2019piqareasoningphysicalcommonsense}

\item {OpenBookQA (OBQA)}: Open-book science QA: combines a provided set of core facts with commonsense/world knowledge to answer questions. \citep{mihaylov2018openbookqa}

\item {RACE}: Multiple-choice reading comprehension from Chinese middle-school English exams. \citep{lai2017race}

\item {MMLU}: ``Massive Multitask Language Understanding''. Questions spanning STEM, humanities, social sciences, etc. \citep{hendrycks2021measuring}

\item {CommonsenseQA (CSQA)}: Multiple-choice QA requiring commonsense relational knowledge (leveraging ConceptNet relations). \citep{talmor2019commonsenseqa}

\item {BoolQ}: Yes/no questions paired with passages to evaluate reading comprehension and entailment-like inference. \citep{clark2019boolq}

\item {GSM8K}: Grade-school math word problems requiring multi-step arithmetic reasoning. \citep{cobbe2021trainingverifierssolvemath}

\item {HumanEval+}: An augmented version of OpenAI's HumanEval \citep{chen2021evaluatinglargelanguagemodels} with many more unit tests per problem to reduce test fragility and overfitting in code generation evaluation. \citep{liu2023codegeneratedchatgptreally}

\item {MBPP}: ``Mostly Basic Programming Problems.'' Short Python programming tasks solvable by entry-level programmers; includes text spec and example tests. \citep{austin2021programsynthesislargelanguage}


\item {NQ}: ``Natural Questions.'' Real user queries paired with Wikipedia pages. \citep{kwiatkowski2019naturalquestions}

\end{itemize}

\section{Performance measures}

\begin{itemize}

\item For generated answers:

\begin{itemize}

\item pass@1 is the proportion of generated pieces of code that
  produce the expected behavior when executed.

\item em (``exact match'') is the proportion of generated endings of a
  sequence that perfectly match a reference solution.


\end{itemize}

\item For multi-choice based on log probabilities:

\begin{itemize}

\item acc\_completion is the proportion of correct responses when the
  choice is based on the sum of the log probabilities normalized with
  the number of tokens of each possible choices.

\item acc\_char is the same as acc\_completion but normalizes with the
  number of characters.

\item macro\_avg/acc\_char is the average of acc\_char over multiple
  sub-categories of questions.

\end{itemize}

\end{itemize}





\clearpage

\section{Performance during training}\label{sec:perf-during-train}

\medskip

\begin{figure}[ht!] 

\center

\includegraphics[scale=0.3,trim=0 0 0 0]{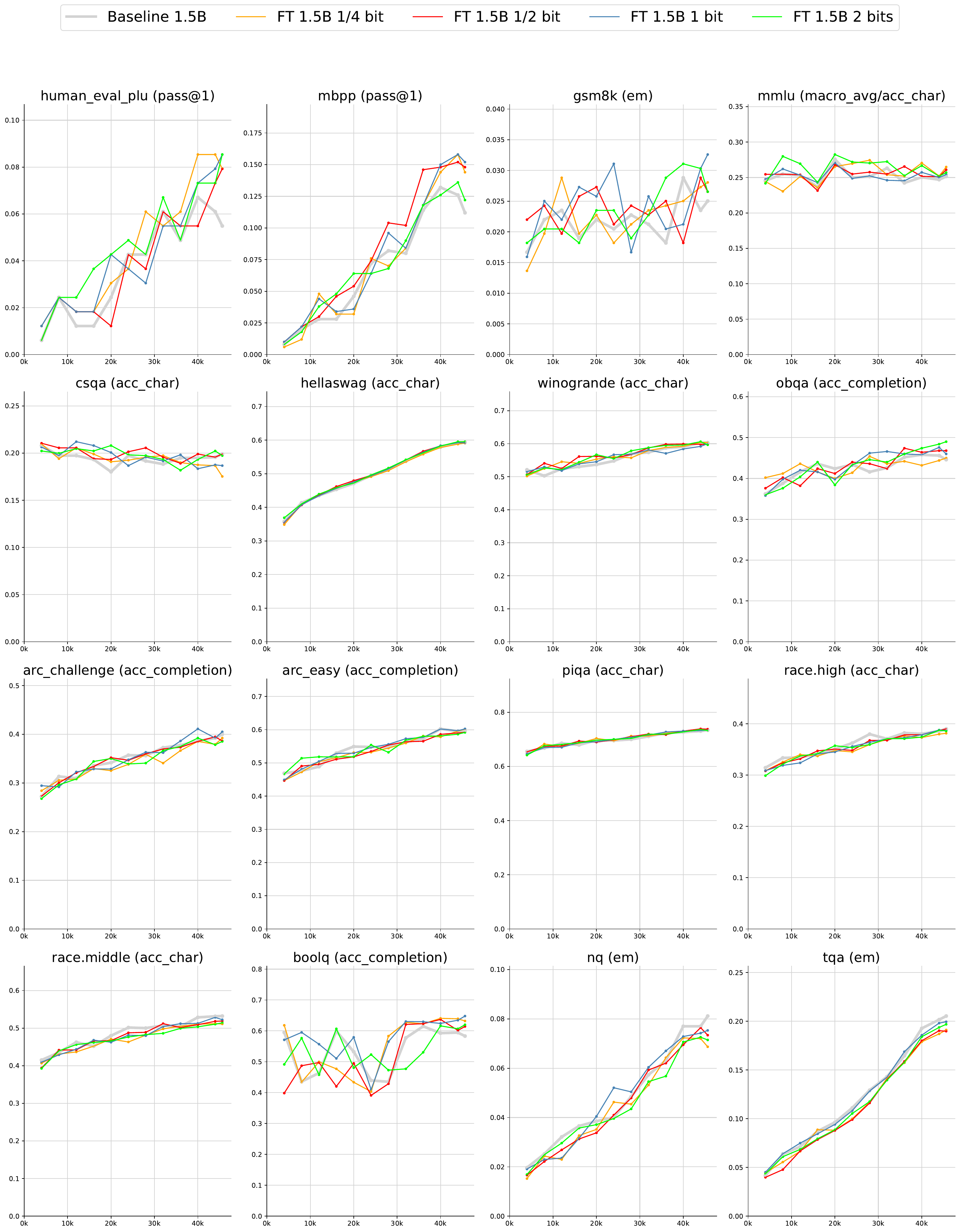}

%

\caption{Experiments with 1.5B models trained on 47B
  tokens. Comparison on standard benchmarks of the baseline and our
  models. The optimization hyperparameters were tuned for the baseline and kept
  unchanged, but the Free Transformers require $3.6\%$ more compute
  and parameters for the encoder.}\label{fig:smodel-perfs}


\end{figure}

\begin{figure}[ht!] 

\center

\includegraphics[scale=0.3,trim=0 0 0 0]{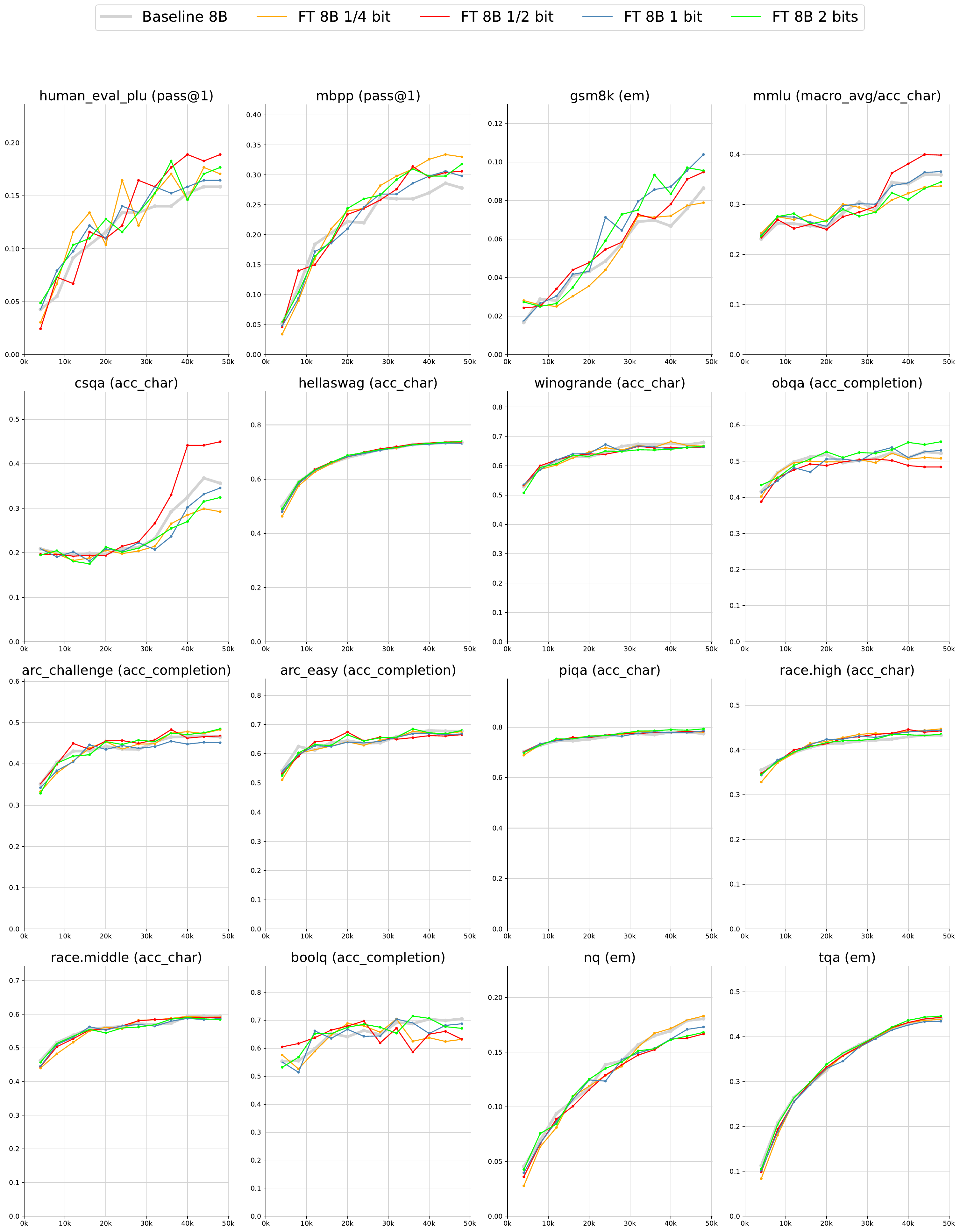}

%
%
\caption{Experiments with 8B models trained on 200B tokens. Comparison
  on standard benchmarks of the baseline and our models. The training
  procedure was tuned for the baseline and kept unchanged, but the
  Free Transformers require $3.1\%$ more compute and parameters for the
  encoder.}\label{fig:llama-perfs}


\end{figure}

\begin{figure}[ht!] 

\center

\includegraphics[scale=0.3,trim=0 0 0 0]{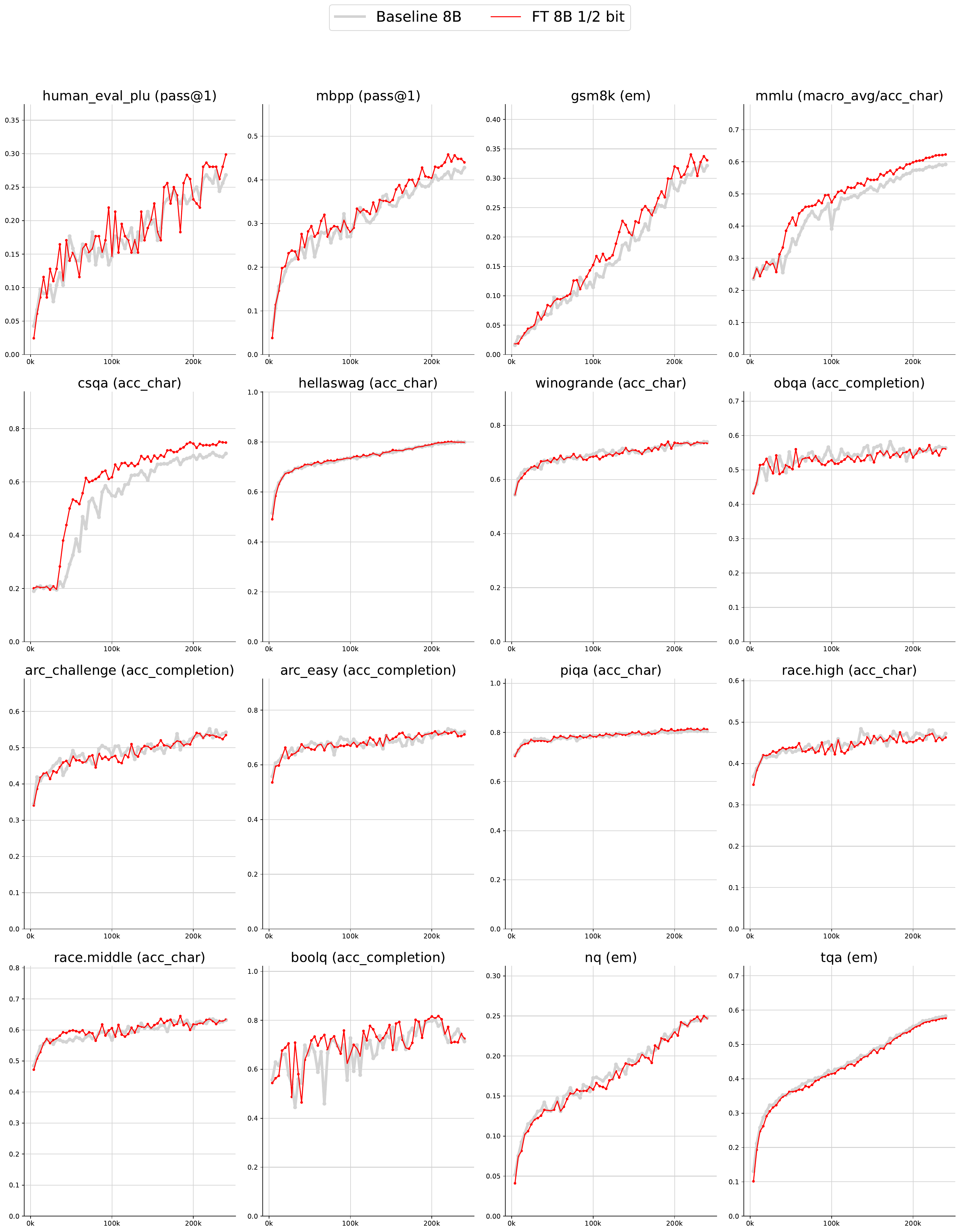}

%
%
\caption{Experiments with 8B models trained on 1T tokens. Comparison
  on standard benchmarks of the baseline and our models. The training
  procedure was tuned for the baseline and kept unchanged, but the
  Free Transformers require $3.1\%$ more compute and parameters for
  the encoder.}\label{fig:llama-perfs-1t}


\end{figure}

\checknbdrafts

\end{document}